\documentclass[twocolumn,letterpaper]{IEEEAerospaceCLS}  

\usepackage{microtype}
\usepackage{graphicx}
\usepackage[dvipsnames,table,xcdraw]{xcolor}
\usepackage{multirow}
\usepackage{subfigure}
\usepackage{booktabs} 
\usepackage{amsmath,amssymb,amsfonts}
\usepackage{bm}
\usepackage{cite}

\begin{document}
\title{Applicability and Challenges of Deep Reinforcement Learning for Satellite Frequency Plan Design}

\author{%
Juan Jose Garau Luis, Edward Crawley, Bruce Cameron\\
Massasachusetts Institute of Technology\\
77 Massachusetts Av. 33-409\\
Cambridge, MA 02139\\
\{garau, crawley, bcameron\}@mit.edu 
\thanks{\footnotesize 978-1-7281-2734-7/20/$\$31.00$ \copyright2020 IEEE}              
}


\maketitle

\thispagestyle{plain}
\pagestyle{plain}

\begin{abstract}


The study and benchmarking of Deep Reinforcement Learning (DRL) models has become a trend in many industries, including aerospace engineering and communications. Recent studies in these fields propose these kinds of models to address certain complex real-time decision-making problems in which classic approaches do not meet time requirements or fail to obtain optimal solutions. While the good performance of DRL models has been proved for specific use cases or scenarios, most studies do not discuss the compromises and generalizability of such models during real operations.

In this paper we explore the tradeoffs of different elements of DRL models and how they might impact the final performance. To that end, we choose the Frequency Plan Design (FPD) problem in the context of multibeam satellite constellations as our use case and propose a DRL model to address it. We identify six different core elements that have a major effect in its performance: the policy, the policy optimizer, the state, action, and reward representations, and the training environment. We analyze different alternatives for each of these elements and characterize their effect. We also use multiple environments to account for different scenarios in which we vary the dimensionality or make the environment non-stationary.

Our findings show that DRL is a potential method to address the FPD problem in real operations, especially because of its speed in decision-making. However, no single DRL model is able to outperform the rest in all scenarios, and the best approach for each of the six core elements depends on the features of the operation environment. While we agree on the potential of DRL to solve future complex problems in the aerospace industry, we also reflect on the importance of designing appropriate models and training procedures, understanding the applicability of such models, and reporting the main performance tradeoffs.

\end{abstract}

\tableofcontents

\section{Introduction}

In order to address the upcoming complex challenges in aerospace and communications research, recent studies are looking into Machine Learning (ML) methods as potential problem-solvers \cite{Abbas2015}. One such method is Deep Reinforcement Learning (DRL), which has had a wide adoption in the community \cite{Luong2019,Ferreira2019}. In the specific case of satellite communications, DRL has already shown its usefulness in problems like channel allocation \cite{Hu2018}, beam-hopping \cite{Hu2020}, and dynamic power allocation \cite{luis2019deep,Zhang2020}.

The motivation behind the use of DRL-based decision-making agents for satellite communications originates mainly from the forthcoming need to automate the control of a large number of satellite variables and beams in real-time. Large flexible constellations, with thousands of beams and a broad range of different users, will need to autonomously reallocate resources such as bandwidth or power in real-time in order to address a highly-fluctuating demand \cite{NorthernSkyResearch2019}. However, the new time and dimensionality requirements pose a considerable challenge to previously-adopted optimization approaches such as mathematical programming \cite{HengWang2013} or metaheuristics \cite{Aravanis2015, Cocco2018, Durand2017}. In contrast, due to the training frameworks common in ML algorithms, DRL has the potential to meet these operational constraints \cite{Luis2020}.

Despite the positive results of DRL for satellite communications and many other fields, the majority of applied research studies mostly focus on best-case scenarios, provide little insight on modelling decisions, and fail to address the operational considerations of deploying DRL-based systems in real-world environments. In addition to the inherent reproducibility problems of DRL algorithms \cite{Henderson2018}, recent studies show the drastic consequences of not properly addressing phenomena present in real-world environments \cite{Dulac-Arnold2020} such as non-stationarity, high-dimensionality, partial-observability, or safety constraints. The lack of robustness against these real-world phenomena lies in part in the modelling design choices, which are a form of inductive biases for DRL models \cite{Hessel2019}. Design choices are rarely discussed in literature and drive a substantial part of the performance, as shown in \cite{Reda2020} for the case of locomotion learning. 

As mentioned, high-dimensionality and non-stationarity are especially important in satellite communications. Mega constellations are already a reality, and they require models that adapt to orders of magnitude of hundreds to thousands of beams. Studies like \cite{Hu2018,Hu2020,Zhang2020} validate the proposed DRL models for cases with less than 50 beams, but obviate the performance in high-dimensional scenarios. In addition, user bases are becoming volatile and their demand highly-fluctuant. Relying on models that assume static user distributions could be detrimental during real-time operations. Therefore, it is crucial that applied research studies also discuss how their models behave against user non-stationarity and how they propose to address any negative influences.

\subsection{Paper goals}

One of the challenges of upcoming large constellations is how to efficiently assign a part of the frequency spectrum to each beam while respecting interference constraints present in the system. This is known as the Frequency Plan Design (FPD) problem. Since users' behavior is highly-dynamic, new frequency plans must be computed in real time to satisfy the demand.  Given these runtime and adaptability requirements, in this paper we propose looking into a DRL-based solution able to make the frequency allocation decisions required at every moment. 

In our study we shift the target from the problem to the implementation methodology. Our aim is to provide a holistic view of how the different elements in a DRL model affect the performance and how their relevance changes depending on the scenario considered. In addition to nominal conditions, we also study the effect of dimensionality and non-stationarity on our models. Our goal is to contribute to the interpretability of DRL models in order to make progress towards real-world deployments that guarantee a robust performance.

\subsection{Paper structure}

The remainder of this paper is structured as follows: Section 2 presents a short overview on DRL and its main performance drivers; Section 3 introduces the FPD problem, focusing on its decisions and constraints; Section 4 outlines our approach to solve the problem with a DRL method; Section 5 discusses the results of the model in multiple scenarios; and finally Section 6 outlines the conclusions of the paper.


\section{Deep Reinforcement Learning}
\label{sec:DRL}

Deep Reinforcement Learning is a Machine Learning subfield that combines the Reinforcement Learning (RL) paradigm \cite{sutton2018reinforcement} and Deep Learning frameworks \cite{goodfellow2016deep} to address complex sequential decision-making problems. The RL paradigm involves the interaction between an \textit{agent} and an \textit{environment} that follows a Markov Decision Process, and is external to the agent. This interaction is sequential and divided into a series of \textit{timesteps}. At each of these timesteps, the agent observes the \textit{state} of the environment and takes an \textit{action} based on it. As a consequence of that action, the agent receives a \textit{reward} that quantifies the value of the action according to the high-level goal of the interaction (e.g., flying a drone or winning a game). Then, the environment updates its state taking into account the action from the agent.

The interaction goes on until a \textit{terminal state} is reached (e.g., the drone lands). In this state, the agent can not take any further action. The sequence of timesteps that lead to a terminal state is known as an \textit{episode}. Based on the experience from multiple episodes, the objective of the agent is to optimize its \textit{policy}, which defines the mapping from states to actions. The agent's optimal policy is the policy that maximizes the expected average or cumulative reward over an episode. This policy might be stochastic depending on the nature of the environment. Table \ref{tab:rlparams} summarizes the parameters defined so far and includes the symbols generally used in literature.

\begin{table}[t]
\renewcommand{\arraystretch}{1.15}
\centering
\caption{Principal Reinforcement Learning parameters and symbols commonly used in literature.}
\label{tab:rlparams}
\begin{tabular}{c|l}
\hline
\rowcolor[HTML]{EFEFEF} 
\textbf{Symbol} & \multicolumn{1}{c}{\cellcolor[HTML]{EFEFEF}\textbf{Parameter}} \\ \hline
$s_t$           & State of the environment at timestep $t$                        \\ 
$a_t$           & Action taken by the agent at timestep $t$                       \\ 
$r_t$           & Reward received at timestep $t$                                 \\ 
$\mathcal{S}$   & Set of possible states \\ 
$\mathcal{A}$   & Set of possible actions \\  
$\mathcal{A}$$(s_t)$ & Set of available actions at state $s_t$ \\ 
$s_T$           & Terminal state \\ 
$\pi$           & Policy \\ 
$\pi^*$         & Optimal policy \\ 
$\pi(a_t|s_t)$  & Probability of action $a_t$ given state $s_t$ \\ \hline
\end{tabular}
\end{table}

In simple environments, the most effective policy is usually given by a table that maps each unique state to an action or probability distribution over actions. However, when the state and/or action spaces are large or continuous, using a tabular policy becomes impractical. Deep Reinforcement Learning (DRL) addresses such cases by substituting tabular policies for neural network-based policies that approximate the mapping from states to actions. Those policies are then updated following Deep Learning algorithms, such as Stochastic Gradient Descent and backpropagation. DRL has shown significant results in a wide variety of areas like games \cite{mnih2015human}, molecular design \cite{Popova2018}, or internet recommendation systems \cite{Theocharous2015}.

There are six elements that jointly drive the performance of a DRL model:
\begin{enumerate}
    \item \textbf{State representation.} We want to look for state representations that capture as much information of the environment as possible and can be easily fed into a neural network. In the case of vision-based robots, these might simply be RGB images from cameras. In other contexts the process sometimes involves more complex state design strategies. It can also be referred as state space.
    \item \textbf{Action space.} It encodes how many different actions the agent can take. Although flexibility matters, reducing the action space is generally beneficial to the algorithm. While, for instance, videogames' actions are straightforward, other environments might require changes to the action space, such as discretization or pruning. It can also be referred as action representation.
    \item \textbf{Reward function.} The reward function usually takes a state and action as inputs and is relevant during the training stage of the algorithm. It should be informative with respect to the final goal of the agent. There might be more than one reward function that correctly guides the learning process, although there is usually a performance gap between classes of strategies: constant reward (i.e., non-zero reward every timestep), sparse reward (i.e., non-zero reward only in terminal states), rollout policy-based reward (i.e., compute reward based on an estimation of the terminal state), etc. 
    \item \textbf{Policy.} In DRL the policy is mainly represented by the neural network architecture chosen. DRL policies commonly include fully-connected layers in addition to convolutional and/or recurrent layers. Each of these layer classes includes multiple subclassifications to consider.
    \item \textbf{Policy optimization algorithm.} There are many algorithms that rely on completely different design choices. In model-free DRL \cite{sutton2018reinforcement}, alternatives include policy gradient methods, Q-learning, or hybrid approaches. Some algorithms might involve the use of additional neural networks to compute performance and value estimates.
    \item \textbf{Training procedure.} The training dataset and training environment constitute an important part of the performance. We want the dataset to reflect all phenomena that might take place during testing and make sure that the training procedure guarantees a robust performance. Sometimes, however, the opportunities of interaction with real environments are limited and models must resort to data-efficient strategies.
\end{enumerate}

\section{Frequency Plan Design Problem}

The Frequency Plan Design problem consists in the assignment of a portion of the available spectrum to every beam in a multibeam satellite system, with the goal of satisfactorily serving all the system's users. Although it is a well-studied problem, the high-dimensionality and flexible nature of new satellites add an additional layer of complexity that motivates the exploration of new algorithmic solutions such as DRL to address it.

\begin{figure}[t]
\centering
\includegraphics[width=.99\linewidth]{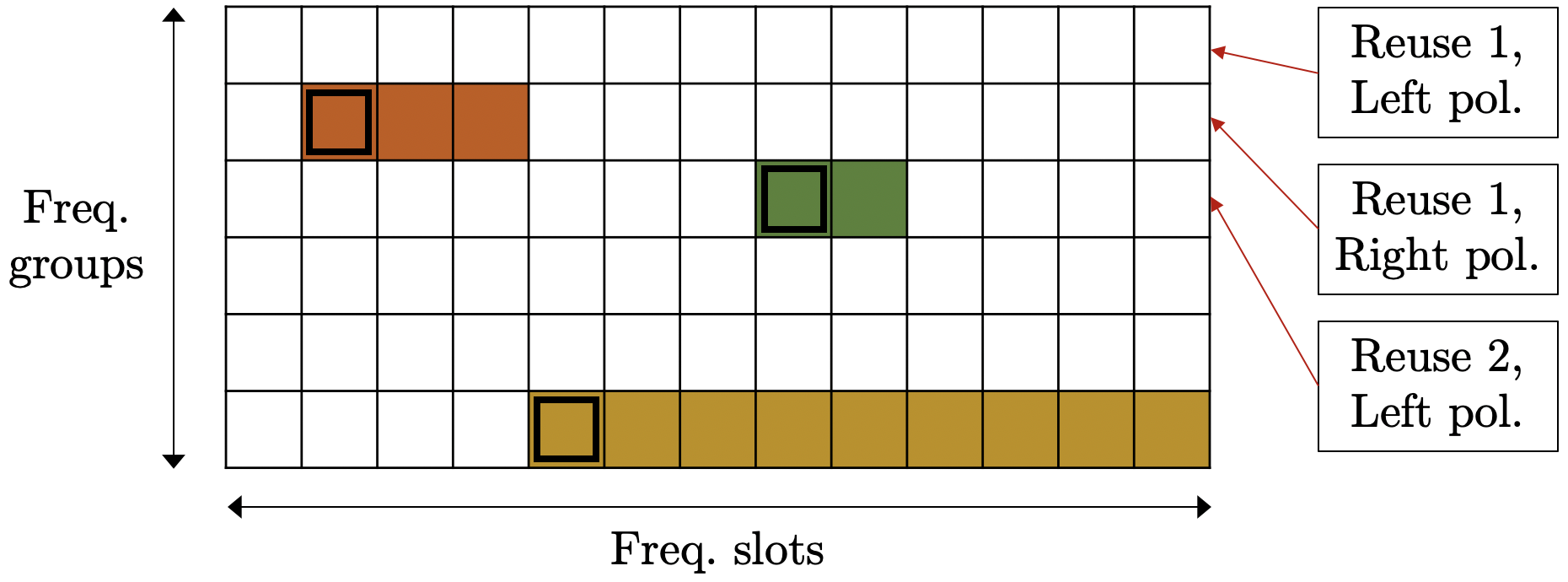}
\caption{Example of the decision-making space for the case $N_{FG}=6$ (frequency groups) and $N_{FS}=13$ (frequency slots). Black squares represent the assignment decision for each beam; colored cells indicate the complete frequency assignment for three different beams.}
\label{fig:grid}
\end{figure}

\subsection{Decisions}

In this paper we consider a multibeam satellite constellation with $N_S$ satellites. We assume a beam placement is provided; we define the total number of beams as $N_B$. Each of these beams has an individual data rate requirement. To satisfy this demand, spectrum resources must be allocated to each beam. For the purpose of this paper, we consider the amount of needed bandwidth for every beam is given, although our framework could be adapted to satisfy dynamic bandwidth requirements given its short runtime. While bandwidth is a continuous resource, most satellites divide up their frequency spectrum into frequency chunks or slots of equal bandwidth, and therefore we consider that each beam needs a certain \textit{discrete} number of \textit{frequency slots}. We denote the number of slots that beam $b$ needs as $BW_b$. The only remaining decisions are therefore to assign which specific frequency slots (among the total available) and the appropriate frequency reuse mechanisms.

In this work we assume satellites can reutilize spectrum. Each satellite of the constellation has an equal available spectrum consisting of $N_{FS}$ consecutive frequency slots. In addition, there is access to frequency reuse mechanisms in the form of frequency reuses and two polarizations. A combination of a specific frequency reuse and polarization is coined as \textit{frequency group}. The constellation has a total of $N_{FG}$ frequency groups, which is twice the amount of reuses (we consider left-handed and right-handed circular polarization available for each beam). For each frequency group, there is an equal number $N_{FS}$ of frequency slots available. 

To better represent this problem and decision space, we use a \textit{grid} in which the columns and the rows represent the $N_{FS}$ frequency slots and the $N_{FG}$ frequency groups, respectively. This is pictured in Figure \ref{fig:grid}. In the case of the frequency groups, we assume these are sorted first, by frequency reuse, and then, by polarization (e.g., row 1 corresponds to reuse 1 and left-handed polarization, and row 2 corresponds to reuse 1 and right-handed polarization). 

The frequency assignment operation for beam $b$ consists of first, selecting one of the $N_{FG}$ frequency groups, and secondly, picking $BW_b$ consecutive frequency slots from the $N_{FS}$ total in that group. In other words, two decisions need to be made per beam: 1) the frequency group and 2) which is the first frequency slot this beam will occupy in the group. This means selecting a specific cell in the grid, which is represented in Figure \ref{fig:grid} by the black squares that designate the assignment decision for each beam. 

\subsection{Constraints}

We identify two types of constraints in this problem. On one hand, since we do not assume a specific orbit, handover operations might occur and we need to account for them if that is the case. This entails that some beams assigned to the same frequency group cannot overlap in the frequency domain because they are powered from the same satellite at some point in time. We define this type of constraints as \textit{intra-group} constraints.

\begin{figure*}[t]
\centering
\includegraphics[width=.9\linewidth]{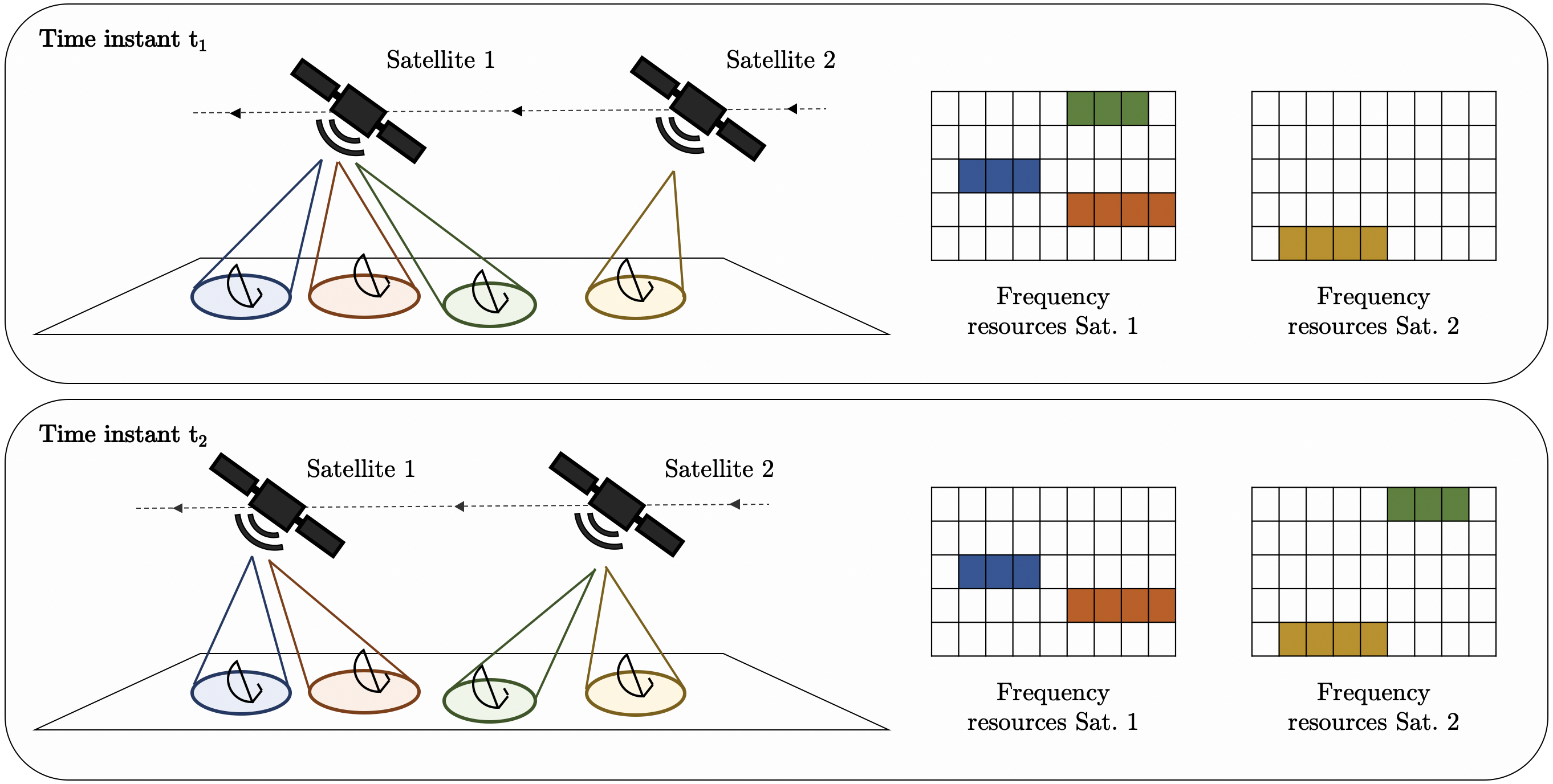}
\caption{Handover operation example. Green beam occupies slots 6, 7, and 8 in group 1 from satellite 1 at time instant $t_1$. At time instant $t_2$, this beam undergoes a handover and uses the same resources from satellite 2, freeing those from the first satellite in the process.}
\label{fig:handover}
\end{figure*}

To better understand intra-group constraints and following the example in Figure \ref{fig:handover}, let's assume beam $b$ (green beam in the figure) is powered from satellite 1 at time instant $t_1$. This beam is assigned to frequency group 1 and slot 6, and needs a bandwidth of 3 slots -- it therefore occupies slots 6, 7, and 8 in group 1. At time instant $t_2$, this beam undergoes a handover operation from satellite 1 to satellite 2. When this occurs, the beam is assigned the same group 1 and slot 6 in satellite 2 and frees them in satellite 1. While we could change the frequency assignment during the handover, this is not always possible due to various factors. As a consequence, a safe strategy is to make sure that, at the moment beam $b$ switches from satellite 1 to satellite 2, slots 6, 7, and 8 in group 1 are available in satellite 2 as well. 


\begin{figure}[t]
\centering
\includegraphics[width=.95\linewidth]{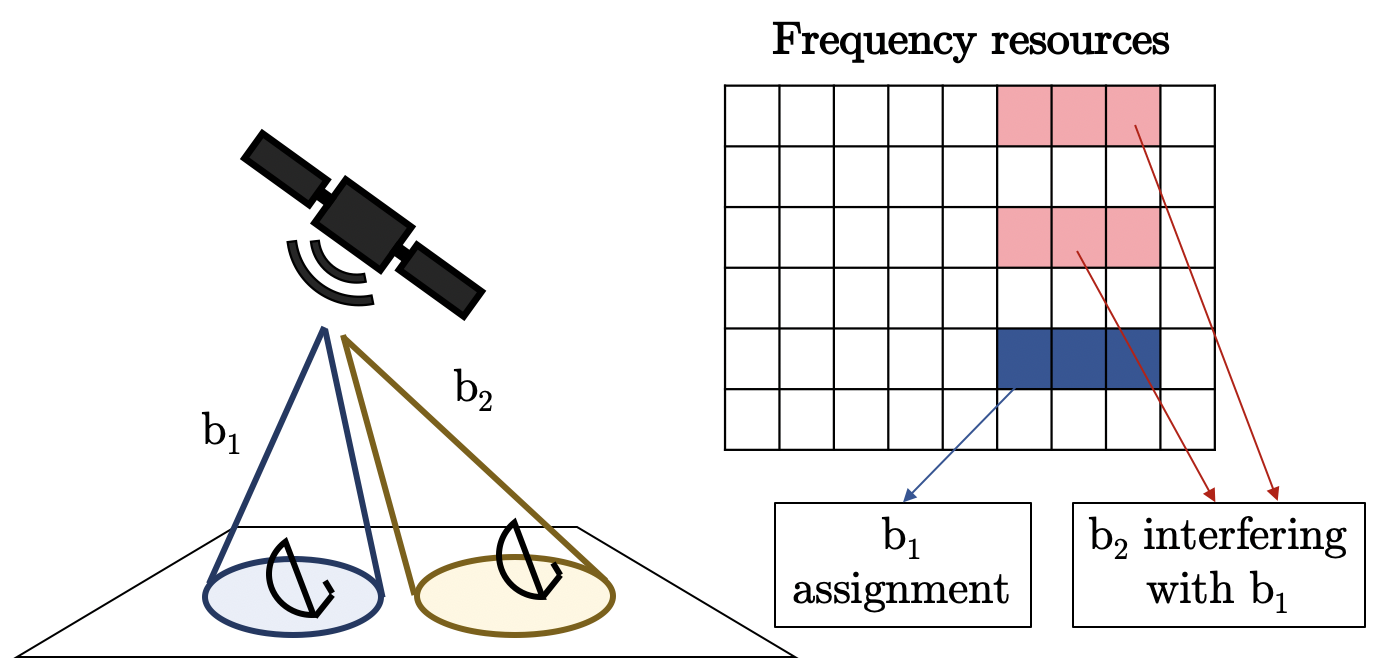}
\caption{Inter-group constraint example between beam $b_1$ and $b_2$. Diagram shows the moment $b_2$ is to be assigned and $b_1$ has already been assigned.}
\label{fig:inter}
\end{figure}

On the other hand, we define the other type of restrictions as \textit{inter-group} constraints. These relate to the cases in which beams that point to close locations might negatively interfere with each other. This situation is more restrictive, as those beams not only must not overlap if they share the same frequency group but also if they share the same polarization. For example, if beams $b_1$ and $b_2$ had an inter-group constraint and beam $b_1$ occupied slots 6, 7, and 8 in group 5, then, following our previous numbering convention, $b_2$ could occupy any slot but slots 6, 7, and 8 in groups 1, 3, 5, 7, etc. This interaction is represented in Figure \ref{fig:inter}. Throughout the remainder of this paper we assume all constraint pairs are known in advance and provided to the model.

\section{Methods}
\label{sec:algos}

The objective of this paper is to design and implement a DRL-based agent capable of making frequency assignment decisions for all $N_B$ beams of a constellation, to evaluate its performance, and to understand its modelling tradeoffs. While there exist algorithms that could address all of the decisions simultaneously, this approach makes the problem challenging for dynamic cases with large values for $N_B$, $N_{FG}$, and $N_{FS}$. The total number of possible different plans is $(N_{FG}N_{FS})^{N_B}$, which poses a combinatorial problem that does not scale well.

As an alternative, we take a sequential decision-making approach in which, following the DRL terminology previously introduced, a \textbf{timestep} is defined as the frequency assignment for just one beam. We then define an \textbf{episode} as the complete frequency assignment of all $N_B$ beams. By doing this, the dimensionality of the problem reduces to $N_{FG}N_{FS}$ for each timestep.

The next step in our formulation involves the definition of the problem-specific elements present in a Reinforcement Learning setup, i.e., state, action, and reward. Generally, the application of DRL to domain-specific problems involves the empirical study of which representation for each element better suits the problem. However, it might be the case that no single representation outperforms the rest in all scenarios, and therefore the representation should be regarded as an additional hyperparameter that depends on the environment conditions. Most domain-specific DRL studies leave these considerations out of their analyses. To emphasize the importance of the representation selection, in this section we propose alternative representations for each element and later study how they affect the overall performance of the model. 

\begin{figure*}[t]
\centering
\includegraphics[width=.99\linewidth]{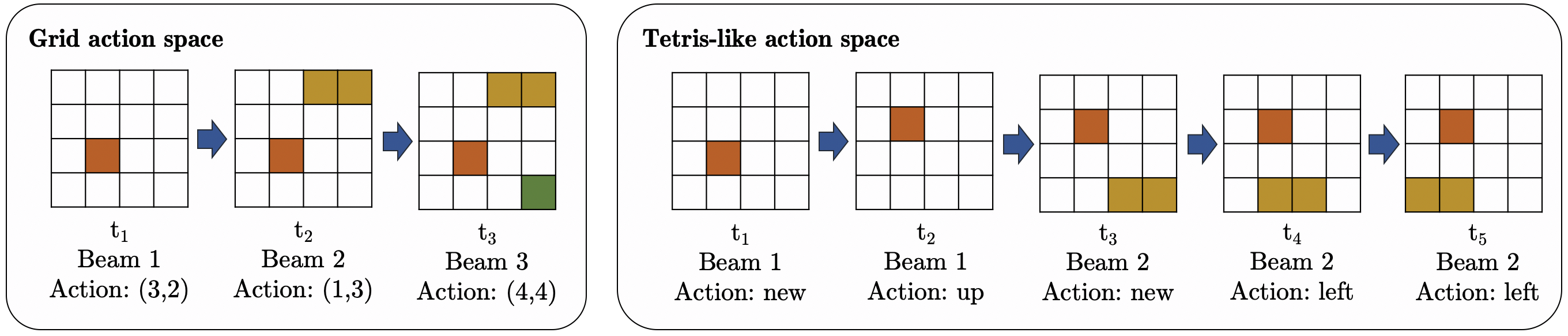}
\caption{Two action spaces considered: \textit{grid} and \textit{tetris-like}.}
\label{fig:action}
\end{figure*}

Regarding the \textbf{action}, two different action spaces are studied. These are pictured in Figure \ref{fig:action}, which shows a scenario/grid with $N_{FG} = 4$ and $N_{FS} = 4$. First, we consider directly choosing a cell in the grid as the action. This action space, which we define as \textit{grid}, consists of $N_{FG}N_{FS}$ different actions. The second action space is defined as \textit{tetris-like} and only contemplates five possible actions. In this space, a random frequency assignment is first made for a beam, i.e., we randonmly choose a cell in the grid. Then, the agent is able to move it up, down, left, and right across the grid until the action \textit{new} is chosen and a new beam undergoes the same procedure. Note that with this latter approach, episodes take a longer and random number of timesteps, since one beam can take more than one timestep to be assigned and there is no restriction on how many intermediate actions should be taken before taking action \textit{new}. The advantage of this representation is the substantial reduction of the action space, which benefits the learning algorithm.

\begin{figure*}[t]
\centering
\includegraphics[width=.95\linewidth]{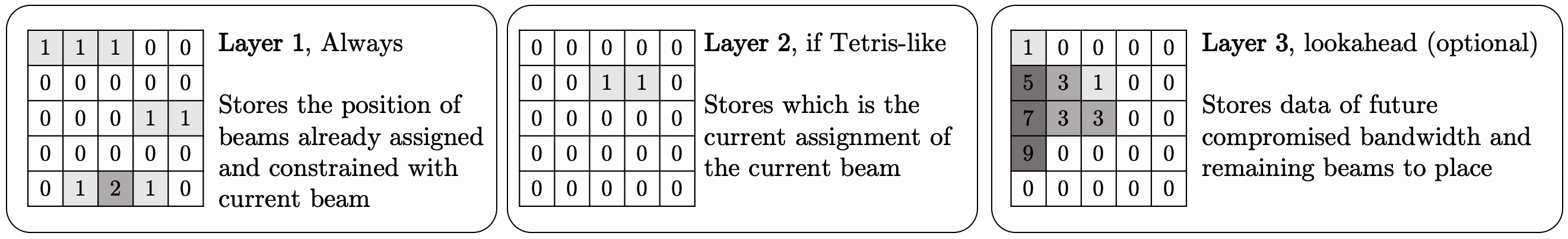}
\caption{Possible layers of the state space.}
\label{fig:state}
\end{figure*}

Next, we also consider two possible representations for the \textbf{state} space. In both of them, the state is defined as a 3-dimensional tensor in which the first dimension size is 1, 2, or 3, and $N_{FG}$ and $N_{FS}$ are the sizes of the second and third dimensions, respectively. We refer to a slice of that tensor along the first dimension as a \textit{layer}. To better understand both representations, we consider the context of a certain timestep, in which we are making the assignment for one specific beam $b_k$, we have already assigned $k-1$ beams, and $N_B-k$ beams remain to be assigned. Below is a description of each layer:
\begin{enumerate}
    \item In both representations the first layer (with dimensions $N_{FG} \times N_{FS}$) stores which grid cells conflict with beam $b_k$, since they are ``occupied'' by at least one of the beams, among the $k-1$ already assigned, that have some kind of constraint with $b_k$. 
    \item In the case the action space is \textit{tetris-like}, the second layer stores the current assignment of beam $b_k$ -- this is done regardless of the state representation chosen. 
    \item The last layer serves as a \textit{lookahead} layer and is optional. This layer contains information regarding the remaining beams, such as the number $N_B - k$, or the amount of bandwidth that will be compromised in the future for beam $b_k$ due to the beams remaining to be assigned which have a intra-group or inter-group constraint with $b_k$. 
\end{enumerate}
Figure \ref{fig:state} shows an example of what these three layers can look like. Whether the last -- lookahead -- layer is included in the tensor or not defines the two state representations considered. We run simulations \textit{with} and \textit{without} the lookahead layer.

Note that our selection of state and action representations entails an additional benefit: we could start from incomplete frequency plans. There might be situations in which we do not want to entirely reconfigure a plan but just reallocate a reduced set of beams -- due to moving users, handover reconfiguration, etc. Furthermore, adding or removing beams to the constellation would not be a problem either. In that sense, the model would be robust against changes in the beam placement. 

Lastly, three alternative functions are considered to define the \textbf{reward}. For the three of them, if the action space is \textit{tetris-like}, the reward function is only applied whenever the action \textit{new} is chosen, otherwise the value $\frac{-1}{N_{FG}\cdot N_{FS}}$ is given as a reward. This is done to avoid the agent finding local optima that consist in just moving a specific beam around without calling the action \textit{new}.

\begin{figure*}[t]
\centering
\includegraphics[width=.99\linewidth]{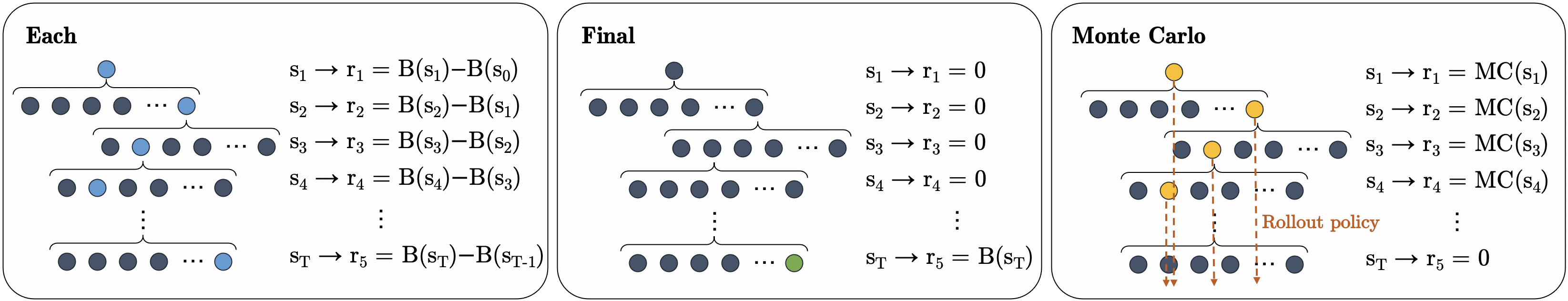}
\caption{Potential reward functions. $B(s_t)$ corresponds to the number of successfully already-assigned beams at state $s_t$.}
\label{fig:reward}
\end{figure*}

To help defining the three reward functions, we denote by $B(s_t)$ the number of successfully already-assigned beams at state $s_t$. We consider a beam to be successfully assigned if it does not violate any constraint. The rewards are then computed as:
\begin{enumerate}
    \item Once a beam is assigned (i.e., any action is taken in the \textit{grid} action space or the action \textit{new} is taken in the \textit{tetris-like} action space), the first reward function consists of computing the difference in the number of successfully already-assigned beams between the states at the current timestep and previous timestep, i.e., $r_t = B(s_t) - B(s_{t-1})$.
    \item The second alternative is to only compute the final quantity of beams that are successfully assigned once the final state $s_T$ is reached and give a reward of zero at all timesteps before that. Therefore, $r_t = B(s_t)$ if $s_t = s_T$, otherwise $0$.
    \item  Finally, inspired by its use in other RL papers \cite{Silver2017}, the third reward function is Monte Carlo-based \cite{sutton2018reinforcement} and uses a rollout policy that randomly assigns the remaining beams at each timestep. This is only done to compute the reward, the outcome of the rollout policy is not taken as agent's actions. The reward equation is $r_t = MC(s_t) = B(s_T)|_{\text{rollout}} - B(s_t)$. Note that $r_T = MC(s_T) = 0$.
\end{enumerate}
We define each of these strategies as \textit{each}, \textit{final}, and \textit{Monte-Carlo} (MC), respectively. Figure \ref{fig:reward} shows a visual comparison between the three options considered.

Before discussing the results, in Table \ref{tab:variations} we summarize the variations that we consider for each of the six elements introduced in Section \ref{sec:DRL}: state, action, reward, policy, policy optimization algorithm, and training procedure. In this section we have described the variations for the first three, in the next section we address the remaining elements, since they specifically relate to the learning framework.

\begin{table}[t]
\renewcommand{\arraystretch}{1.15}
\centering
\caption{Principal DRL parameters and the variations we test in this paper.}
\label{tab:variations}
\begin{tabular}{c|l}
\hline
\rowcolor[HTML]{EFEFEF} 
\textbf{DRL Element}  & \multicolumn{1}{c}{\cellcolor[HTML]{EFEFEF}\textbf{Variations considered}} \\ \hline
State repr.  & \textit{Lookahead} and \textit{without} lookahead                                     \\ 
Action repr. & \textit{Grid} and \textit{tetris-like}                                                                         \\ 
Reward                   & \textit{Each}, \textit{final}, and \textit{Monte-Carlo}                                                                         \\ 
Policy                   & CNN+MLP and CNN+LSTM                                                                         \\ 
Policy Opt.                   & DQN and PPO                                                                         \\
Training proc.                   & Same as test and harder than test                                                                         \\ \hline
\end{tabular}
\end{table}

\section{Results}
\label{sec:results}

To test the presented framework, we first simulate a scenario with 100 beams, 7 satellites, $N_{FG} = 4$, and $N_{FS} = 20$. From a set of 5,000 beams based on real data, we randomly create a train and a test dataset, disjoint with respect to each other. For each episode of the training stage, we randomly select 100 beams from the train dataset. Then, during the test stage we randomly select 100 beams from the test dataset and evaluate the policy on those beams. 

Since the nature of the problem is discrete, we initially select \textit{Deep Q-Network} \cite{mnih2015human} as the \textbf{policy optimization algorithm} and later in this section we compare it against a policy gradient algorithm. We use 8 different environments in parallel, this way we can share the experience throughout the training stage and have statistics over the results during the test stage. 

We start using a \textbf{policy} consisting of a Convolutional Neural Network (CNN) with 2 layers (first layer with 64 5x5 filters, second layer with 128 3x3 filters), then 2 fully-connected layers (first layer with 512 units, second layer with 256 units), and an output layer. The fully-connected layers can be commonly referred as Multi Layer Perceptron (MLP). ReLU activation units and normalization layers are used in all cases.

\subsection{Full enumeration analysis}

We do a full enumeration analysis and test each possible combination of action-state-reward representation for a total of 50k timesteps. At this point, we also focus on the \textbf{training procedure} and test which improvements there might be if we train the agent to do the assignment with 200 beams instead of 100, to make the task harder and to potentially obtain a more robust policy. In this latter case, the test evaluation is still carried out on 100 beams. In addition, we extend the full enumeration search to include the \textit{discount factor} $\gamma$ \cite{sutton2018reinforcement}, a DRL hyperparameter, and evaluate three possible values: 0.1, 0.5, and 0.9. A larger discount factor implies that the agent takes into account longer term effects of its actions, whereas lower discount factors are related to greedier policies. 

\begin{table*}[t]
\renewcommand{\arraystretch}{1.15}
\centering
\caption{Average number of successfully-assigned beams out of 100 in the test data. A random policy achieves 83.5.}
\label{tab:test}
\begin{tabular}{
>{\columncolor[HTML]{EFEFEF}}l |
>{\columncolor[HTML]{EFEFEF}}c |
>{\columncolor[HTML]{FFFFFF}}c 
>{\columncolor[HTML]{FFFFFF}}c 
>{\columncolor[HTML]{FFFFFF}}c 
>{\columncolor[HTML]{FFFFFF}}c |
>{\columncolor[HTML]{FFFFFF}}c 
>{\columncolor[HTML]{FFFFFF}}c 
>{\columncolor[HTML]{FFFFFF}}c 
>{\columncolor[HTML]{FFFFFF}}c }
\hline
{\color[HTML]{000000} \textbf{}}       & \multicolumn{1}{l|}{\cellcolor[HTML]{EFEFEF}{\color[HTML]{000000} \textbf{}}} & \multicolumn{4}{c|}{\cellcolor[HTML]{C0C0C0}{\color[HTML]{000000} \textbf{100-beam training}}}                                                                                                                                                                                                                                 & \multicolumn{4}{c}{\cellcolor[HTML]{C0C0C0}{\color[HTML]{000000} \textbf{200-beam training}}}                                                                                                                                                                                                                                  \\ \cline{3-10} 
{\color[HTML]{000000} \textbf{}}       & {\color[HTML]{000000} \textbf{}}                                              & \multicolumn{2}{c|}{\cellcolor[HTML]{D0D0D0}{\color[HTML]{000000} \textbf{Grid}}}                                                                                        & \multicolumn{2}{c|}{\cellcolor[HTML]{D0D0D0}{\color[HTML]{000000} \textbf{Tetris-like}}}                                                            & \multicolumn{2}{c|}{\cellcolor[HTML]{D0D0D0}{\color[HTML]{000000} \textbf{Grid}}}                                                                                        & \multicolumn{2}{c}{\cellcolor[HTML]{D0D0D0}{\color[HTML]{000000} \textbf{Tetris-like}}}                                                             \\ \cline{3-10} 
{\color[HTML]{000000} \textbf{Reward}} & {\color[HTML]{000000} \textbf{$\gamma$}}                                      & \multicolumn{1}{c|}{\cellcolor[HTML]{E0E0E0}{\color[HTML]{000000} \textbf{Lookahead}}} & \multicolumn{1}{c|}{\cellcolor[HTML]{E0E0E0}{\color[HTML]{000000} \textbf{Without}}} & \multicolumn{1}{c|}{\cellcolor[HTML]{E0E0E0}{\color[HTML]{000000} \textbf{Lookahead}}} & \cellcolor[HTML]{E0E0E0}{\color[HTML]{000000} \textbf{Without}} & \multicolumn{1}{c|}{\cellcolor[HTML]{E0E0E0}{\color[HTML]{000000} \textbf{Lookahead}}} & \multicolumn{1}{c|}{\cellcolor[HTML]{E0E0E0}{\color[HTML]{000000} \textbf{Without}}} & \multicolumn{1}{c|}{\cellcolor[HTML]{E0E0E0}{\color[HTML]{000000} \textbf{Lookahead}}} & \cellcolor[HTML]{E0E0E0}{\color[HTML]{000000} \textbf{Without}} \\ \hline
{\color[HTML]{000000} \textbf{}}       & {\color[HTML]{000000} \textbf{0.1}}                                           & {\color[HTML]{000000} \textbf{99.8}}                                                       & {\color[HTML]{000000} 95.6}                                                          & {\color[HTML]{000000} 92.0}                                                       & {\color[HTML]{000000} 95.8}                                     & {\color[HTML]{000000} 98.9}                                                       & {\color[HTML]{000000} 97.9}                                                          & {\color[HTML]{000000} 87.5}                                                       & {\color[HTML]{000000} 98.2}                                     \\
{\color[HTML]{000000} \textbf{Each}}   & {\color[HTML]{000000} \textbf{0.5}}                                           & {\color[HTML]{000000} 98.9}                                                       & {\color[HTML]{000000} 96.2}                                                          & {\color[HTML]{000000} 90.5}                                                       & {\color[HTML]{000000} 96.6}                                     & {\color[HTML]{000000} 97.6}                                                       & {\color[HTML]{000000} 94.4}                                                          & {\color[HTML]{000000} 80.8}                                                       & {\color[HTML]{000000} 90.9}                                     \\
{\color[HTML]{000000} \textbf{}}       & {\color[HTML]{000000} \textbf{0.9}}                                           & {\color[HTML]{000000} 96.6}                                                       & {\color[HTML]{000000} 93.1}                                                          & {\color[HTML]{000000} 81.5}                                                       & {\color[HTML]{000000} 66.9}                                     & {\color[HTML]{000000} 95.2}                                                       & {\color[HTML]{000000} 93.4}                                                          & {\color[HTML]{000000} 70.2}                                                       & {\color[HTML]{000000} 66.6}                                     \\ \hline
{\color[HTML]{000000} \textbf{}}       & {\color[HTML]{000000} \textbf{0.1}}                                           & {\color[HTML]{000000} 47.5}                                                       & {\color[HTML]{000000} 53.9}                                                          & {\color[HTML]{000000} 68.8}                                                       & {\color[HTML]{000000} 67.4}                                     & {\color[HTML]{000000} 23.8}                                                       & {\color[HTML]{000000} 20.9}                                                          & {\color[HTML]{000000} 52.5}                                                       & {\color[HTML]{000000} 45.5}                                     \\
{\color[HTML]{000000} \textbf{Final}}  & {\color[HTML]{000000} \textbf{0.5}}                                           & {\color[HTML]{000000} 46.6}                                                       & {\color[HTML]{000000} 56.0}                                                          & {\color[HTML]{000000} 39.9}                                                       & {\color[HTML]{000000} 70.1}                                     & {\color[HTML]{000000} 21.1}                                                       & {\color[HTML]{000000} 21.2}                                                          & {\color[HTML]{000000} 67.0}                                                       & {\color[HTML]{000000} 69.9}                                     \\
{\color[HTML]{000000} \textbf{}}       & {\color[HTML]{000000} \textbf{0.9}}                                           & {\color[HTML]{000000} 54.9}                                                       & {\color[HTML]{000000} 67.2}                                                          & {\color[HTML]{000000} 71.9}                                                       & {\color[HTML]{000000} 63.6}                                     & {\color[HTML]{000000} 25.8}                                                       & {\color[HTML]{000000} 22.6}                                                          & {\color[HTML]{000000} 60.0}                                                       & {\color[HTML]{000000} 67.1}                                     \\ \hline
{\color[HTML]{000000} \textbf{}}       & {\color[HTML]{000000} \textbf{0.1}}                                           & {\color[HTML]{000000} 19.5}                                                       & {\color[HTML]{000000} 17.6}                                                          & {\color[HTML]{000000} 45.2}                                                       & {\color[HTML]{000000} 45.2}                                     & {\color[HTML]{000000} 2.5}                                                        & {\color[HTML]{000000} 2.9}                                                           & {\color[HTML]{000000} 50.9}                                                       & {\color[HTML]{000000} 34}                                       \\
{\color[HTML]{000000} \textbf{MC}}     & {\color[HTML]{000000} \textbf{0.5}}                                           & {\color[HTML]{000000} 9.1}                                                        & {\color[HTML]{000000} 3.0}                                                           & {\color[HTML]{000000} 67.9}                                                       & {\color[HTML]{000000} 36.1}                                     & {\color[HTML]{000000} 2.6}                                                        & {\color[HTML]{000000} 1.9}                                                           & {\color[HTML]{000000} 41.1}                                                       & {\color[HTML]{000000} 20.2}                                     \\
{\color[HTML]{000000} \textbf{}}       & {\color[HTML]{000000} \textbf{0.9}}                                           & {\color[HTML]{000000} 2.2}                                                        & {\color[HTML]{000000} 2.6}                                                           & {\color[HTML]{000000} 37.8}                                                       & {\color[HTML]{000000} 47.5}                                     & {\color[HTML]{000000} 2.2}                                                        & {\color[HTML]{000000} 3.2}                                                           & {\color[HTML]{000000} 34.6}                                                       & {\color[HTML]{000000} 24.1}                                     \\ \hline
\end{tabular}
\end{table*}

The results of the enumeration analysis for the 72 different simulations are shown in Table \ref{tab:test}, which contains the average number of successfully-assigned beams during test time (out of 100) for each combination of action, state, reward, discount factor $\gamma$, and number of training beams. The average is computed across the 8 parallel environments. For reference, we compare these results against a totally random policy, which achieves an average of 83.5 successfully-assigned beams.

It is observed that using the reward strategy \textit{each} leads to better outcomes, and using the \textit{grid} action space with the state space using \textit{lookahead} does better on average. For our case, there is no apparent advantage in using reward strategies that delay the reward until the end of the episode or rely on rollout policies; individual timesteps provide enough information to guide the policy optimization. To better understand the impact of the modelling decisions, we also do several significance tests and further analyze the tradeoffs:

\begin{itemize}
    \item There is no advantage in training with 200 beams instead of 100 (p-value = 0.08). Therefore, training with more beams does not make a more robust policy for our case. This is even less impactful when $\gamma=0.1$ (p = 0.90), since the policy behaves greedily regardless of how many beams remain to be assigned. The strategy for placing the first 100 beams is similar both cases. 
    \item The discount factor affects the performance of the policy (p = 10$^{-8}$), but there is no significant performance difference when using $\gamma = 0.1$ or $\gamma =0.5$ (p = 0.83). The performance worsens when $\gamma = 0.9$. The learned policy relies more on a greedy behavior to be successful, which is consistent with other methods that have addressed the problem \cite{PachlerdelaOsa2020}. 
    \item With $\gamma\leq 0.5$, the \textit{each} reward strategy, and \textit{grid} action space, the state space affects the performance (p = 10$^{-6}$). It is better to include the lookahead layer.
    \item With $\gamma\leq 0.5$, the \textit{each} reward strategy, and \textit{tetris-like} action space, the state space affects the performance (p = 10$^{-3}$). Not including the lookahead layer (\textit{without} in the table) in the state achieves better results in this case.
    \item With $\gamma\leq 0.5$ and the \textit{each} reward strategy, using the \textit{grid} state space is a better alternative than using \textit{tetris-like} (p = 10$^{-5}$), since it supports more flexibility to make an assignment.
    \item Overall, the \textit{tetris-like} action space appears to be less sensitive to the reward strategy chosen, especially when comparing across the simulations using the \textit{Monte-Carlo} reward strategy.
\end{itemize}

The lookahead layer plays an important role together with the \textit{grid} action space, since the agent has total freedom to place a beam, and therefore being informed of the remaining assignments is beneficial. In contrast, in the case of the \textit{tetris-like} action space, the agent is limited by the initial random placement of the beam, and therefore the information of what is to come is not as useful. This is an example of how different representations can interact with each other.

\subsection{Scalability analysis}

Given the high-dimensionality of future satellite systems, we also analyze the impact scalability has on our framework and run new simulations for 500 beams, 7 satellites, $N_{FG} = 16$, $N_{FS} = 80$, $\gamma=0.1$, and the \textit{each} reward strategy, and compare all 4 combinations for the state and action representations. In this case we train each model for a total of 200k timesteps. These results are shown in Table \ref{tab:test2} and can be compared against a totally random policy, which achieves 429.9 successfully-assigned beams.

The main conclusion of this analysis is that when the dimensionality of the problem increases, we can actually achieve a better outcome using the \textit{tetris-like} action space, as opposed to the 100-beam case. The \textit{grid} action space does even worse than random, which is due to the large amount of actions (1,280 for this case) and an inapprpriate exploration-exploitation balance. Overall, there is no impact in terms of the state representation in this example (p = 0.64), but if we focus on the \textit{tetris-like} action space, there is (p = 0.001); we still do better \textit{without} the lookahead layer. These results prove that relying on a single representation for a specific problem might not be a robust strategy during operation. Understanding the limitations of specific representations is essential in order to make these models deployable and, more importantly, reliable. 

\begin{table}[t]
\renewcommand{\arraystretch}{1.15}
\centering
\caption{Average number of successfully-assigned beams out of 500 in the test data using the \textit{each} reward function and $\gamma=0.1$. A random policy achieves 429.9.}
\label{tab:test2}
\begin{tabular}{c|c}
\hline
\rowcolor[HTML]{EFEFEF} 
\textbf{\begin{tabular}[c]{@{}c@{}}Action and state \\ representation\end{tabular}} & \textbf{\begin{tabular}[c]{@{}c@{}}Number of successfully-\\ assigned beams\end{tabular}} \\ \hline
Grid and Lookahead                                                                       & 320.1                                                                                     \\
Grid and Without                                                                    & 328.1                                                                                     \\
Tetris-like and Lookahead                                                                & 461.6                                                                                     \\ 
Tetris-like and Without                                                             & \textbf{478.9}                                                                                     \\ \hline
\end{tabular}
\end{table}

\subsection{Policy and Policy Optimization Algorithm}

Two of the key elements of a DRL model listed in Section \ref{sec:DRL} and Table \ref{tab:variations}, the policy and the policy optimization algorithm, remain to be studied for our problem. These heavily rely on the progress made by the Deep Learning and Reinforcement Learning research communities. To address them, we now extend the analyses to compare the results of the Deep Q-Network (DQN) algorithm, and the CNN+MLP policy, both used in all simulations so far, against other alternatives. 

In the case of the \textbf{policy optimization algorithm}, we choose the Proximal Policy Optimization (PPO) \cite{schulman2017proximal} method, a \textit{Policy Gradient} and \textit{on-policy} algorithm \cite{sutton2018reinforcement}, to be compared against DQN \cite{mnih2015human}, a \textit{Q-learning} and \textit{off-policy} algorithm. PPO optimizes the policy on an end-to-end fashion, by doing gradient ascent on its parameters according to a function of the cumulative reward over an episode. Additionally, it clips the gradients to avoid drastic changes to the policy. DQN focuses on optimizing the prediction of the \textit{value} of taking a certain action in a given state; it then uses these predictions to choose an action. Also, it stores all the agent's experience and makes use of it over time by ``replaying'' past episodes and training on them. We refer to the original papers for a full description of each method. We use OpenAI's baselines \cite{baselines} to implement each method.

Regarding the \textbf{policy} network, we substitute the MLP for a 256-unit Long Short-Term Memory (LSTM) \cite{Hochreiter1997} network, which belongs to the recurrent networks class. We compare a policy constituted by CNN+MLP against a CNN+LSTM one to evaluate the impact of using a hidden state in our formulation. Recurrent neural networks take advantage of temporal dependencies in the data and therefore can potentially perform better in sequential problems.

Table \ref{tab:test3} shows the results of using both policies and both policy optimization algorithms for 4 different cases: the 100-beam scenario using the \textit{grid} action space with both state representations, the 500-beam scenario using the \textit{tetris-like} action space with both state representations, and two additional 1,000-beam and 2,000-beam scenarios for the \textit{tetris-like} action space \textit{without} using the lookahead layer in the state representation. For each case the random policy successfully assigns 83.5, 429.9, 799.7, and 1,137.7 beams on average, respectively. Since we have concluded the \textit{tetris-like} action space better suits high-dimensional scenarios, we explore its performance in the thousand-beam range as well. In all cases, the \textit{each} reward strategy and $\gamma = 0.1$ is used. All simulations belonging to the same scenario are trained for an equal number of timesteps.

\begin{table}[t]
\renewcommand{\arraystretch}{1.15}
\centering
\caption{Average number of successfully-assigned beams in the test data when comparing different combinations of policy and policy optimization algorithm. The \textit{each} reward function and $\gamma=0.1$ are used in all cases.}
\label{tab:test3}
\begin{tabular}{c|c|c|c}
\hline
\rowcolor[HTML]{EFEFEF} 
\textbf{\begin{tabular}[c]{@{}c@{}}Action and state \\ representation\end{tabular}} & \textbf{\begin{tabular}[c]{@{}c@{}}DQN\\ MLP\end{tabular}} & \textbf{\begin{tabular}[c]{@{}c@{}}PPO \\ MLP\end{tabular}} & \textbf{\begin{tabular}[c]{@{}c@{}}PPO  \\ LSTM\end{tabular}} \\ \hline

\rowcolor[HTML]{E0E0E0} 
\multicolumn{4}{c}{\cellcolor[HTML]{E0E0E0}100 beams, Random: 83.5} \\
\rowcolor[HTML]{E0E0E0}
\multicolumn{4}{c}{\cellcolor[HTML]{E0E0E0}$N_{FG}=4$, $N_{FS}=20$}    
\\ \hline

Grid, Lookahead                                                                       & \textbf{99.8}                               & 94.9    & 94.9                           \\ 
Grid, Without                                                                    & 95.6                                 & 97.5     & 91.5                         \\ \hline
\rowcolor[HTML]{E0E0E0} 
\multicolumn{4}{c}{\cellcolor[HTML]{E0E0E0}500 beams, Random: 429.9}     \\  
\rowcolor[HTML]{E0E0E0} 
\multicolumn{4}{c}{\cellcolor[HTML]{E0E0E0}$N_{FG}=16$, $N_{FS}=80$} \\ \hline
Tetris-like, Lookahead                                                                & 461.6                                 & 460.5        & 470.0                      \\
Tetris-like, Without                                                             & \textbf{478.9}                                 & 478.0      & 465.1                         \\ \hline
\multicolumn{4}{c}{\cellcolor[HTML]{E0E0E0}1,000 beams, Random: 799.7} \\
\multicolumn{4}{c}{\cellcolor[HTML]{E0E0E0}$N_{FG}=20$, $N_{FS}=100$}
\\ \hline

Tetris-like, Without & \textbf{962.3} & 936.0 & 957.0 \\ \hline

\multicolumn{4}{c}{\cellcolor[HTML]{E0E0E0}2,000 beams, Random: 1,137.7} \\
\multicolumn{4}{c}{\cellcolor[HTML]{E0E0E0}$N_{FG}=20$, $N_{FS}=100$}
\\ \hline

Tetris-like, Without & \textbf{1,746.0} & 1,139.0 & 967.0 \\ \hline

\end{tabular}
\end{table}

In the scenarios with 1,000 beams or less, we can observe there is no significant advantage in using PPO over DQN, although DQN consistently outperforms PPO in each scenario. Looking at the 2,000-beam scenario, however, the performance difference emerges. Note that the 1,000-beam and 2,000-beam simulations use the same values for $N_{FG}$ and $N_{FS}$. Therefore, in the 2,000-beam case, the performance worsening occurs during the assignment of the last 1,000 beams. During the last timesteps of such a high-dimensional scenario, having a prediction over multiple actions, as DQN does, proves to be a better approach, as opposed to PPO's method, which relies on the single action that the policy provides.  

\begin{figure*}[t]
\centering
\includegraphics[width=.99\linewidth]{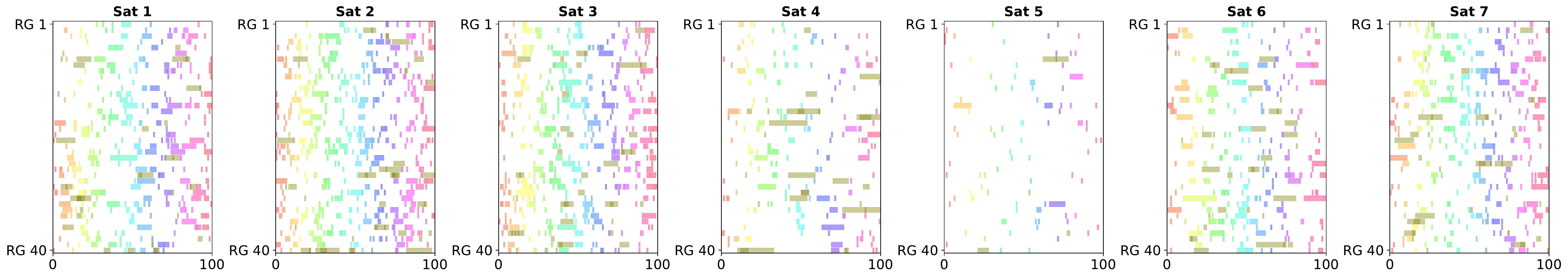}
\caption{2,000-beam Frequency Plan example obtained by the DQN-based agent using the CNN+MLP policy, the \textit{tetris-like} action space, the state representation \textit{without} the lookahead layer, the \textit{each} reward function, and $\gamma=0.1$. Golden cells indicate non-successful assignments (i.e., a constraint is violated), any other color indicates a successful assignment. 1,766 beams are successfully-assigned in this case.}
\label{fig:fp_example}
\end{figure*}

The same occurs if we compare the CNN+MLP policy (MLP in the table) against the CNN+LSTM policy (LSTM in the table). Given the greedy behaviour of the policy, having the LSTM's hidden state does not offer any advantage when placing the last 1,000 beams in the 2,000-beam scenario. Although allowing more training iterations could help reducing the performance gap, these results prove that the choices of the policy and the policy optimization algorithm are especially important under certain circumstances. For the FPD problem in the context of megaconstellations, we need to care about the thousand-beam range. To help visualizing the performance of these models, Figure \ref{fig:fp_example} shows the assignment results for one of the 8 environments in the 2,000-beam, DQN, and CNN+MLP scenario. A total of 1,766 beams are successfully assigned.

\subsection{Non-stationarity}

We now address the issue of non-stationarity and its impact on trained DRL models. In the context of the FPD problem, we are interested in the phenomena that might change from training to operation, such as the bandwidth distribution, the number of total beams, or the constraint distribution. We choose the bandwidth distribution for our analyses. Specifically, we study how the performance changes when the test environment includes beams with more average bandwidth demand than those in the training set. We carry out simulations in which the average bandwidth demand per beam in the test set is 2-times and 4-times larger. We do this for the 100-beam scenario and the \textit{grid} action space, as well as for the 500-beam scenario and the \textit{tetris-like} action space. For both cases, we compare both state representations and also run the random policy. The outcome of this analysis can be found in Table \ref{tab:test4}.

\begin{table}[t]
\renewcommand{\arraystretch}{1.15}
\centering
\caption{Average number of successfully-assigned beams in the test data when comparing different imbalances between train and test data. The \textit{each} reward function, the DQN algorithm, the CNN+MLP policy, and $\gamma=0.1$ are used in all cases.}
\label{tab:test4}
\begin{tabular}{c|c|c|c}
\hline
\rowcolor[HTML]{EFEFEF} 
\textbf{\begin{tabular}[c]{@{}l@{}}Action, State\\ representation\end{tabular}} & \textbf{\begin{tabular}[c]{@{}c@{}}Same\\ demand\end{tabular}} & \textbf{\begin{tabular}[c]{@{}c@{}}2-times\\ demand\end{tabular}} & \textbf{\begin{tabular}[c]{@{}c@{}}4-times\\ demand\end{tabular}} \\ \hline
\rowcolor[HTML]{E0E0E0} 
\multicolumn{4}{c}{\cellcolor[HTML]{E0E0E0}100 beams, $N_{FG}=4$, $N_{FS}=20$}     \\  \hline
Grid, Lookahead                                                                      & \textbf{99.8}                                                           & \textbf{84.9}                                                              & 52.6                                                              \\ 
Grid, Without                                                                   & 95.6                                                           & 84.4                                                              & 65.3                                                              \\ 
Random                                                                          & 83.5                                                           & 81.3                                                              & \textbf{71.9}                                                              \\ \hline
\rowcolor[HTML]{E0E0E0} 
\multicolumn{4}{c}{\cellcolor[HTML]{E0E0E0}500 beams, $N_{FG}=16$, $N_{FS}=80$}     \\  \hline
Tetris-like, Lookahead                                                               & 461.6                                                          & \textbf{453.0}                                                             & 249.0                                                             \\ 
Tetris-like, Without                                                            & \textbf{478.9}                                                          & 434.0                                                             & \textbf{337.0}                                                             \\ 
Random                                                                          & 429.9                                                          & 400.0                                                             & 227.0                                                             \\ \hline
\end{tabular}
\end{table}

The results show that, as expected based on other DRL studies \cite{Dulac-Arnold2020}, non-stationarity does negatively affect the performance of trained DRL models for the FPD problem. In our case, it correlates with the average bandwidth demand per beam difference between the train and test data. In the 100-beam scenario, the performance gap between the random policy and the agent is reduced for the 2-times case, and then random performs better in the 4-times case. 

Something similar occurs in the 500-beam scenario, although the random policy does not beat the agent in any case. This is a consequence having a bigger search space and the more frequent use of the actions \textit{up} and \textit{down} in the \textit{tetris-like} action space to make changes to the resource group assignment rather than changing the assignment within the same resource group. If we analyze the actions taken during test time for this latter scenario, the \textit{up} and \textit{down} actions were taken, on average, 2.5 times more than the \textit{left} and \textit{right} actions. Furthermore, non-stationarity affects the usefulness of the lookahead layer for the \textit{tetris-like} action space. In the analysis from Table \ref{tab:test2} we have seen that not using it was a better alternative, whereas in the 2-times demand case the agent does better with it. There is a certain advantage in knowing what is to come in non-stationary scenarios, although there is a limit to how beneficial the lookahead layer is, as observed in the 4-times demand case. 

There are different alternatives to be robust in non-stationary environments. The first involves, after identifying the sources of non-stationarity, devising adequate training datasets that include episodes capturing those sources. In this case, it is essential to use a state representation that successfully incorporates information on the condition of each source. Secondly, we can make use of algorithms that refine the agent's actions under contingency scenarios. These algorithms could guarantee a most robust behavior at the expense of computing time, such as in \cite{garau20a}, where a DRL model is combined with a subsequent Genetic Algorithm to maximize robustness. Finally, we can rely not only on a single agent but on an ensemble of agents that specialize on different tasks and are trained in precisely different ways, such as the method proposed in \cite{Ghosh2017}. In all cases, retraining the models over time from collected experience during operation is a good practice to capture any possible change in the environment.

\subsection{Impact on real-world scenarios}

The proposed models have shown a quantitatively good performance in the majority of the scenarios considered. On average, the agent successfully-assigns 99.8\% of beams in the 100-beam case, 95.8\% in the 500-beam case, 96.2\% in the 1,000-beam case, and 87.3\% in the 2,000-beam case. The main advantage of this approach is that evaluating a neural network is substantially faster than relying on other algorithms such as metaheuristics, while still showing a good performance. This aligns with what most applied DRL studies conclude. To correct not fulfilling the entirety of the constraints, we could have a subsequent ``repairing'' algorithm to address the conflicting beams, which could make up 1 to 15\% of the total based on our numbers.

However, the main consideration behind these numbers is that they correspond to different models that use different state and action representations. Unlike many applied DRL studies, we have regarded representations as additional hyperparameters at every stage of our analyses. Given the importance of high-dimensionality in real-world operations, we have also focused on cases with hundreds to thousands of beams, as opposed to works that exclusively consider cases with less than 100 beams. We could not have extracted our conclusions without making these decisions for our analyses. When studying a specific DRL model with defined hyperparameters, understanding which are the limitations of those hyperparameters and testing the models under real-world conditions is at least as important as, if not more important than achieving or beating the state-of-the-art performance. 

We have also studied the limitations resulting from environment non-stationarity. We have validated the Machine Learning community's findings on the negative effects of non-stationary sources in the environment. It is not common to find this kind of analysis in domain-specific DRL papers. If the goal is make DRL deployable, it is essential that we address contingency cases in our problems, either by capturing the sources of non-stationarity in the training datasets, or by devising strategies to mitigate the impact during real-time operations. Otherwise we fail to meet the requirements of operating in real-world environments and, more importantly, reduce the reliability of our systems.


\section{Conclusions}

In this paper we have addressed the design and implementation tradeoffs of a Deep Reinforcement Learning (DRL) agent to carry out frequency assignment tasks in a multibeam satellite constellation. DRL models are getting attention in the aerospace community mostly due to its fast decision-making and its adaptability to complex non-linear optimization problems. In our work we have chosen the Frequency Plan Design (FPD) problem as a use case and identified six elements that drive the performance of DRL models: the state representation, the action representation, the reward function, the policy, the policy optimization algorithm, and the training strategy. We have defined multiple variations for each of these elements and compared the performance differences in separate scenarios. We have put a special focus on high-dimensionality and non-stationarity, being two of the main phenomena present in the upcoming satellite communications landscape.

The results show that DRL is an adequate method to address the FPD problem, since it successfully assigns 85\% to 99\% of the beams for cases with 100 to 2,000 beams. However, no single state-action combination outperforms the rest for all cases. When the dimensionality of the problem is low, the \textit{grid} action space and the state representation using the \textit{lookahead} layer perform better. In contrast, the \textit{tetris-like} action space \textit{without} the lookahead layer in the state representation are a better option for high-dimensional scenarios. These findings validate our hypothesis that representation should be strongly considered as an additional hyperparameter in applied DRL studies. We have also seen that using the Deep Q-Network algorithm in combination with a convolutional neural network and a fully-connected neural network as the policy works best for all scenarios, being especially advantageous for the 2,000-beam case. Regardless of the scenario, the obtained policy has shown a greedy behavior that benefits from informative rewards at each timestep.

At the end of the paper, we have reflected on different considerations that are usually left out of applied DRL studies. Our analyses on the effect of non-stationarity have helped motivating the discussion. When the average bandwidth demand per beam distribution substantially differs between the train and test data the number of successfully-assigned beams by agent decreases, performing worse than random for some cases. We emphasize the need to identify the potential sources of non-stationarity, understand its potential effects for the DRL model during real operation cycles, and propose solutions to mitigate any negative influences. A compromise between topping performance metrics and characterizing the limitation of the models and its hyperparameters is the only way to advance in the successful deployment of DRL and other Machine Learning-based technologies.







\acknowledgments
This work was supported by SES. The authors would like to thank SES for their input to this paper and their financial support. The simulations are based on a representative example of a MEO constellation system, which was provided by SES \cite{SES}. The authors would also like to thank Iñigo del Portillo and Markus Guerster, for contributing to the initial discussions that motivated this work; and Skylar Eiskowitz and Nils Pachler, for reviewing the manuscript.

\bibliographystyle{IEEEtran}
\bibliography{library}




\thebiography

\begin{biographywithpic}
{Juan Jose Garau Luis}{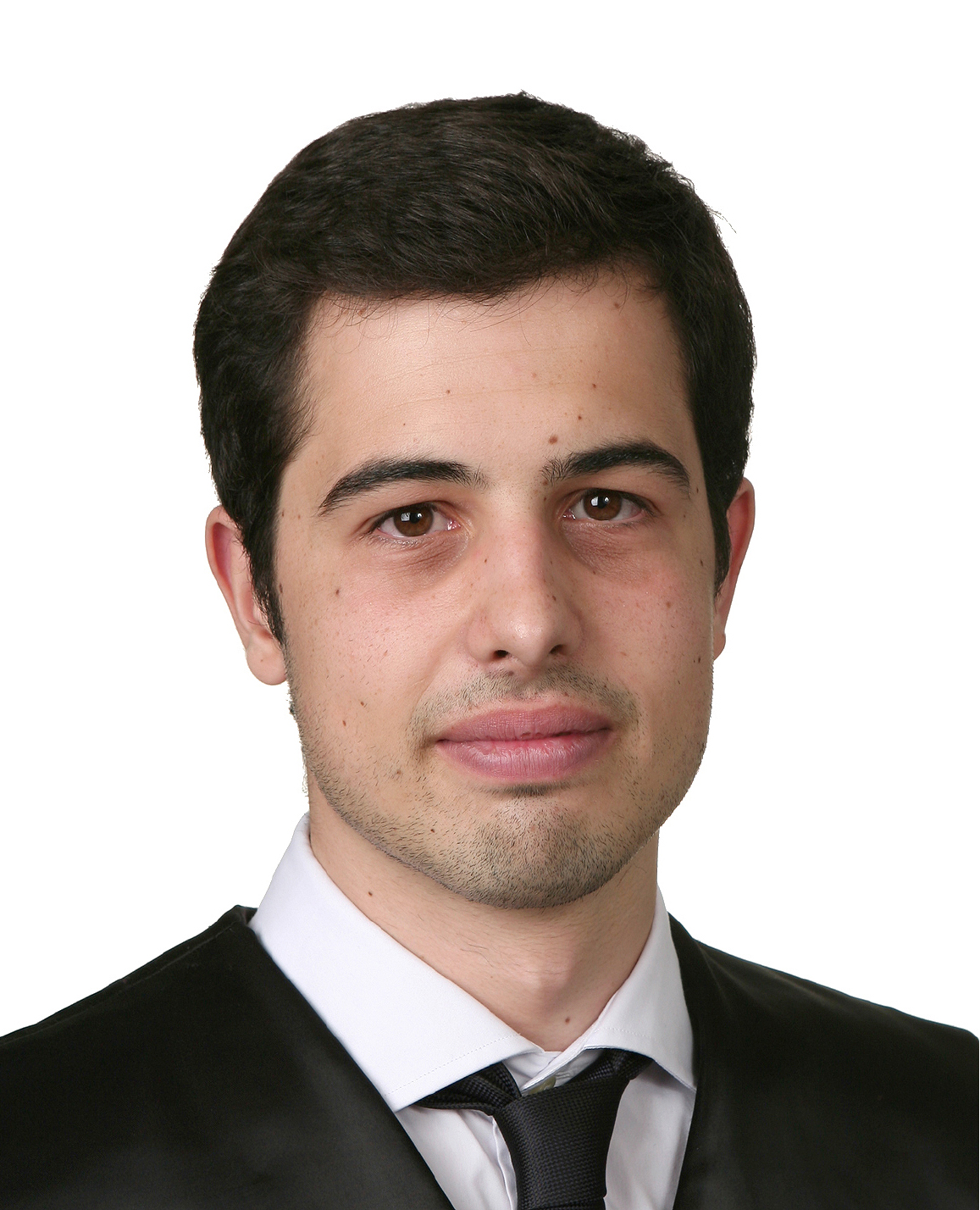}
is a Ph.D. Candidate in the Department of Aeronautics and Astronautics at MIT. In 2017, under the CFIS program, he received two B.Sc. degrees in Telecommunications Engineering and Industrial Engineering from Universitat Politecnica de Catalunya, in Spain. In 2020, he received his M.Sc. degree in Aerospace Engineering from MIT. He has worked as a data scientist and Machine Learning engineer at Novartis, Arcvi, and the Barcelona Supercomputing Center. His research interests include autonomous systems, Machine Learning, and communication networks.
\end{biographywithpic}

\begin{biographywithpic}
{Prof. Edward Crawley}{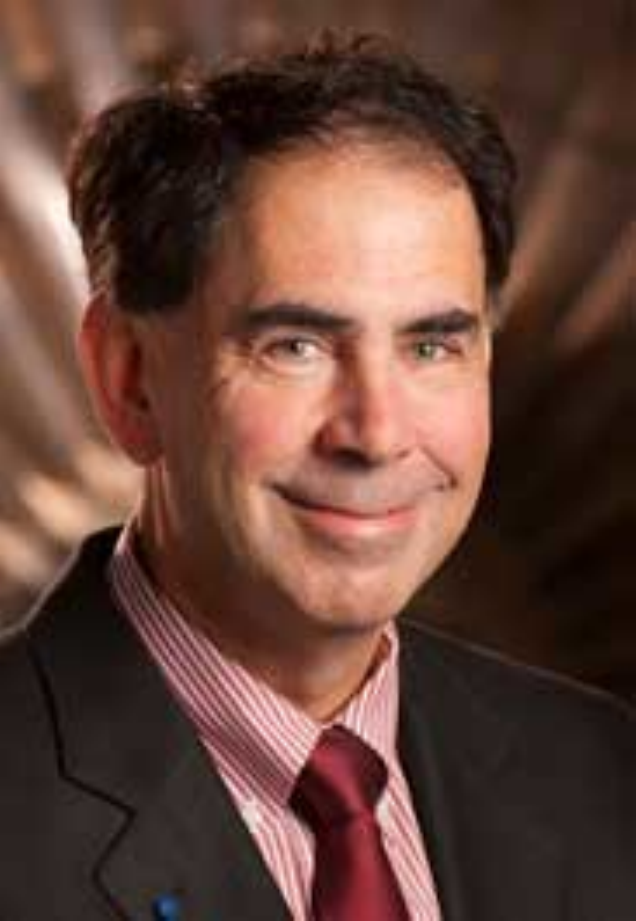}
received an Sc.D. in Aerospace Structures from MIT in 1981. His early research interests centered on structural dynamics, aeroelasticity, and the development of actively controlled and intelligent structures. Recently, Dr. Crawley's research has focused on the domain of the architecture and design of complex systems. From 1996 to 2003 he served as the Department Head of Aeronautics and Astronautics at MIT, leading the strategic realignment of the department. Dr. Crawley is a Fellow of the AIAA and the Royal Aeronautical Society (UK), and is a member of three national academies of engineering. He is the author of numerous journal publications in the AIAA Journal, the ASME Journal, the Journal of Composite Materials, and Acta Astronautica. He received the NASA Public Service Medal. Recently, Prof Crawley was one of the ten members of the presidential committee led by Norman Augustine to study the future of human spaceflight in the US.
\end{biographywithpic}

\begin{biographywithpic}
{Dr. Bruce Cameron}{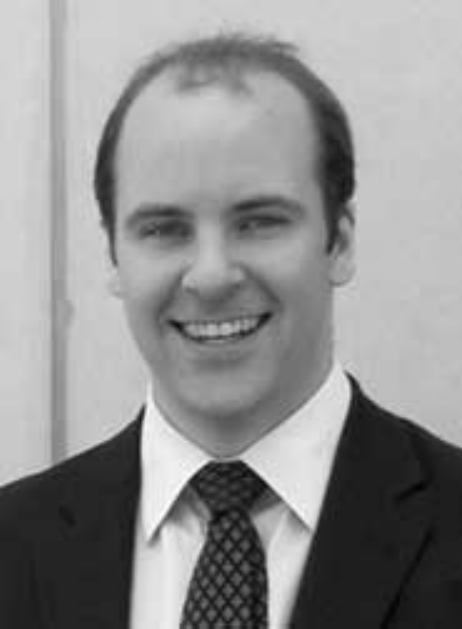}
is a Lecturer in Engineering Systems at MIT and a consultant on platform strategies. At MIT, Dr. Cameron ran the MIT Commonality study, a 16 firm investigation of platforming returns. Dr. Cameron’s current clients include Fortune 500 firms in high tech, aerospace, transportation, and consumer goods. Prior to MIT, Bruce worked as an engagement manager at a management consultancy and as a system engineer at MDA Space Systems, and has built hardware currently in orbit. Dr. Cameron received his undergraduate degree from the University of Toronto, and graduate degrees from MIT.
\end{biographywithpic}

\end{document}